\documentclass[english]{article}
\usepackage[T1]{fontenc}
\usepackage[utf8]{inputenc}
\usepackage{array}
\usepackage{url}
\usepackage{multirow}
\usepackage{amsbsy}
\usepackage{amstext}
\usepackage{amssymb}
\usepackage{graphicx}
\usepackage[numbers]{natbib}
\usepackage{microtype}

\makeatletter

\providecommand{\tabularnewline}{\\}

\usepackage[final]{nips_2018}
\usepackage[pagebackref=true,breaklinks=true,letterpaper=true,colorlinks,bookmarks=false]{hyperref}
\usepackage{url}            % simple URL typesetting
\usepackage{booktabs}       % professional-quality tables
\usepackage{amsfonts}       % blackboard math symbols
\usepackage{nicefrac}       % compact symbols for 1/2, etc.
\usepackage{rotating}
\usepackage{babel}
\makeatother

\usepackage{babel}
\begin{document}
\title{Hybrid Knowledge Routed Modules for Large-scale Object Detection}
\author{Chenhan Jiang$^*$ \\
Sun Yat-Sen University \\
\texttt{jchcyan@gmail.com} \\
\And 
Hang Xu\thanks{Both authors contributed equally to this work.}\\
Huawei Noah's Ark Lab \\
\texttt{xbjxh@live.com} \\ 
\AND 
Xiaodan Liang\thanks{Corresponding Author} \\ 
School of Intelligent Systems Engineering\\
Sun Yat-Sen University \\ 
\texttt{xdliang328@gmail.com} \\
\And 
Liang Lin \\
Sun Yat-Sen University \\
\texttt{linliang@ieee.org} \\}

\maketitle
\vspace{-2mm}

\begin{abstract}
The dominant object detection approaches treat the recognition of
each region separately and overlook crucial semantic correlations
between objects in one scene. This paradigm leads to substantial performance
drop when facing heavy long-tail problems, where very few samples
are available for rare classes and plenty of confusing categories
exists. We exploit diverse human commonsense knowledge for reasoning
over large-scale object categories and reaching semantic coherency
within one image. Particularly, we present Hybrid Knowledge Routed
Modules (HKRM) that incorporates the reasoning routed by two kinds
of knowledge forms: an explicit knowledge module for structured constraints
that are summarized with linguistic knowledge (e.g. shared attributes,
relationships) about concepts; and an implicit knowledge module that
depicts some implicit constraints (e.g. common spatial layouts). By
functioning over a region-to-region graph, both modules can be individualized
and adapted to coordinate with visual patterns in each image, guided
by specific knowledge forms. HKRM are light-weight, general-purpose
and extensible by easily incorporating multiple knowledge to endow
any detection networks the ability of global semantic reasoning. Experiments
on large-scale object detection benchmarks show HKRM obtains around
34.5\% improvement on VisualGenome (1000 categories) and 30.4\% on
ADE in terms of mAP. Codes and trained model can be found in \url{https://github.com/chanyn/HKRM}.
\end{abstract}
\vspace{-2mm}

\section{Introduction\label{sec:Introduction}}

The most state-of-the-art object detection methods \citep{NIPS2009_3766,ren2015faster,dai2016r,cai2017cascade}
follow the region-based paradigm, which treats the classification
and boundingbox regression of each region proposal separately. The
detection performance purely relies on the discriminative capabilities
of region features, which often depends on sufficient training data
for each category. Such paradigm thus obtains substantial performance
drop when dealing with large-scale detection task \citep{torralba2004sharing,hoffman2014lsda}
that recognizes and localizes a large number of categories (e.g. 3000
classes in VG \citep{krishnavisualgenome}). The long-tail problem
is very common, where very few samples exist for rare classes, such
as pepperoni and bagel. On the other hand, detection challenges such
as heavy occlusion, class ambiguities and tiny-size objects become
more severe due to more categories within one image. However, humans
can still identity objects precisely under complex circumstances because
of the remarkable reasoning ability resorting to commonsense knowledge.
This inspires us to explore how to incorporate diverse knowledge forms
into current detection paradigm in a light-weight and effective way,
in order to mimic human reasoning procedure.

\begin{figure}[t]
\begin{centering}
\includegraphics[scale=0.38]{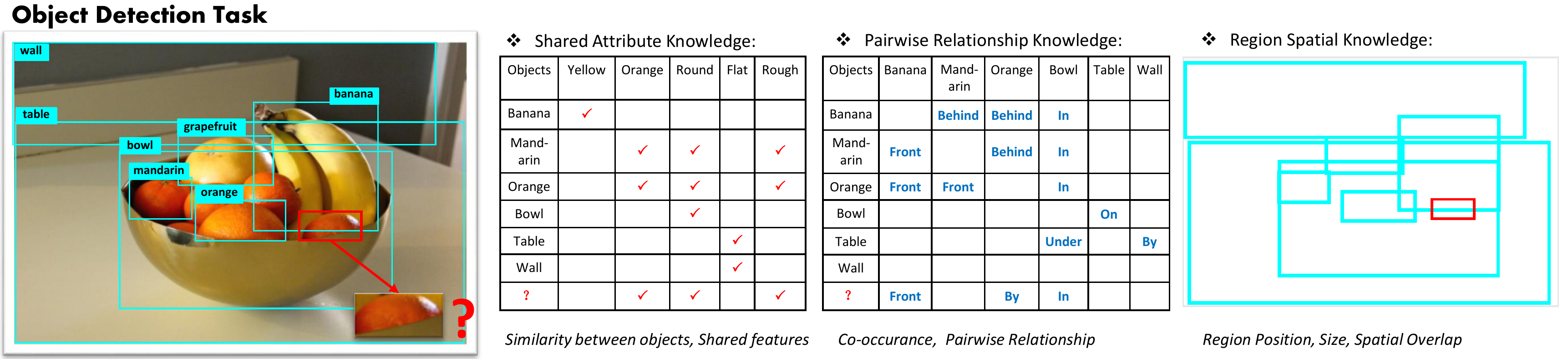}
\par\end{centering}
\vspace{-1mm}

\caption{\label{fig:motivation}An example of how different types of commonsense
knowledge can facilitate large-scale object detection, especially
for rare classes (e.g. the obscured mandarin). We illustrate three
useful  knowledge forms: attribute knowledge, relationship knowledge
and spatial knowledge.}

\vspace{-3mm}
\end{figure}
When humans watch a scene \citep{biederman1982scene}, each object
is not identified individually. Different knowledge obtained by a
human commonsense can help to make a correct identification by considering
global semantic coherency. An example of hybrid knowledge reasoning
in Figure \ref{fig:motivation} would be to identify the obscured
``mandarin'' (bottom-right). Human can recognize it is a mandarin
learned from hybrid commonsense: a) this round object is orange and
just like the other nearby mandarins (shared attribute knowledge);
b) this object is in the bowl (pairwise relationship knowledge); c)
this object has moderate size and its position is near to other fruits
(spatial layout).

Recently, some works incorporate knowledge via direct relation modeling
\citep{mensink2012metric,deng2014large,hu2017relation} or iterative
reasoning architecture \citep{marino2016more,chen2017spatial,chen2018iterative}.
Different from recent implicit relation networks \citep{hu2017relation,Wang2017Non}
that learned inter-region relationships in an implicit and uncontrollable
way, recently an iterative reasoning \citep{chen2018iterative} was
proposed to combine both local and global reasoning. However, they
take only region predictions of a basic detection network, rather
than enhancing intermediate feature representations. Furthermore,
they directly use statistic edge connections in a prior knowledge
graph while ignoring the compatibility of prior knowledge with visual
evidence in each image. Given diverse object appearances and correlations
in each image, personalized edge connections with respect to each
knowledge form should be adaptive for different regions. On the contrary,
our work aims to develop in-place knowledge modules which can not
only explicitly incorporate any kinds of commonsense knowledge (both
explicit or implicit) for better semantic reasoning but also link
external knowledge with visual observations in each image in an adaptive
way.

In this paper, we propose Hybrid Knowledge Routed Modules (HKRM) to
incorporate multiple semantic reasoning routed by two major kinds
of knowledge forms: an explicit knowledge module that exploits structure
constraints that are summarized with linguistic knowledge (e.g. shared
attributes, co-occurrence and relationships), and an implicit knowledge
module to encode some implicit commonsense constraints over object
(e.g. common spatial layouts). Instead of building category-to-category
graph \citep{li2015gated,niepert2016learning,kipf2016semi,marino2016more,dai2017detecting},
each knowledge module in HKRM learns adaptive context connections
for each pair of regions by regarding a specific prior knowledge graph
as external supervisions, rather than fixing the connections. Our
HKRM is general-purposed and extensible by easily integrating several
individualized knowledge modules instantiated with any chosen knowledge
forms to pursue more advanced and hybrid semantic reasoning. As a
showcase, we experiment with three kinds of knowledge forms in this
paper: the attribute knowledge (e.g. color, status), pairwise relationship
knowledge such as co-occurrence and object-verb-subject relationship,
the spatial knowledge including layout, size and overlap. HKRM is
light-weight and easily plugged into any detection network for endowing
its ability in global reasoning.

Our HKRM thus enables sharing visual features among certain regions
with similar attributes, pairwise relationship or spatial relationship.
The recognition and localization of difficult regions with heavy occlusions,
class ambiguities and tiny-size problems can be thus remedied by discovering
adaptive contexts from other regions guided by external knowledge.
Another merit of HKRM lies in the ability of distilling common characteristics
among common/uncommon categories so that the problem of crucial imbalanced
categories can be alleviated.

The proposed HKRM outperforms the state-of-the-art Faster RCNN \citep{ren2015faster}
with a large margin on two large-scale object detection benchmarks,
that is, ADE \citep{zhou2017scene} with 445 object classes and VG
\citep{krishnavisualgenome} with 1000 or 3000 classes. Particularly,
our HKRM achieves around 34.5\% of mAP improvement on VG (1000 categories),
26.5\% on VG (3000 categories) and 30.4\% on ADE. More interestingly,
further analysis shows our HKRM module can provide meaningful explanations
about how different commonsense knowledge can help perform reasonable
visual reasoning and what each module actually learn with the guidance
of external knowledge.

\section{Related Work}

\begin{figure}[t]
\begin{centering}
\includegraphics[scale=0.26]{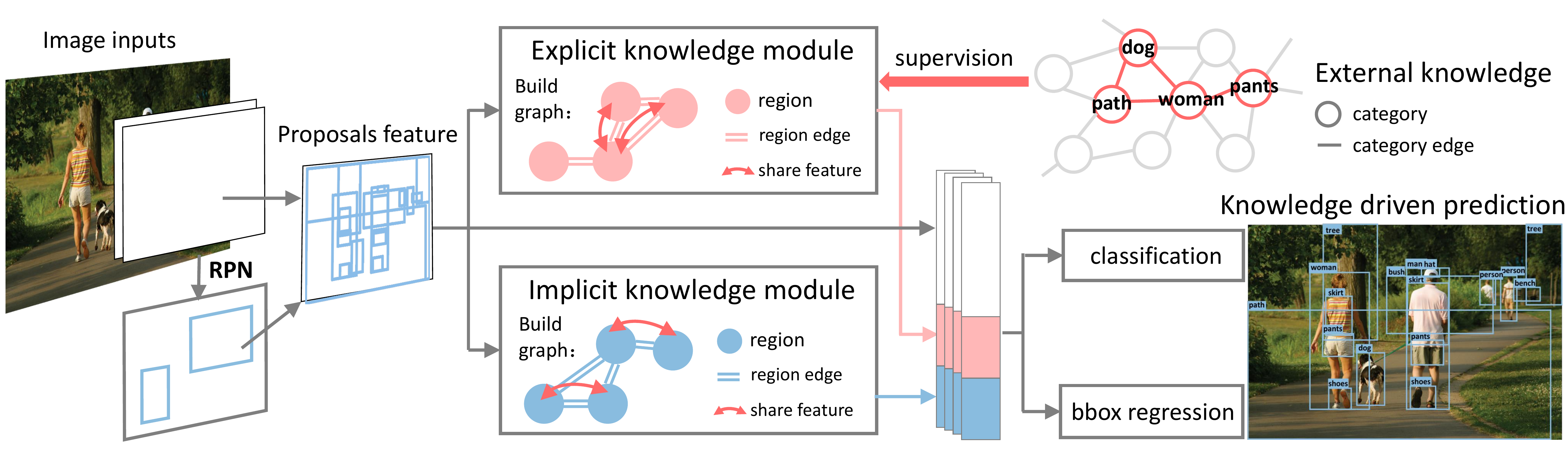}
\par\end{centering}
\vspace{-1mm}

\caption{\label{fig:Overview-of-our_HKRM}Overview of our HKRM, including two
kinds of general modules: an explicit knowledge module to incorporate
external knowledge and an implicit knowledge module to learn knowledge
without explicit definitions or being summarized by human, such as
spatial layouts. An adaptive region-to-region knowledge graph is constructed
by regarding each specified external knowledge as the supervision
of edge connections. The features of each region node are then enhanced
through integrating several individual knowledge modules instantiated
with distinct knowledge forms. The evolved features after each module
are combined to produce final object detection results. }

\vspace{-3mm}
\end{figure}
\textbf{Object Detection. }Big progress has been made recent years
on object detection due to the use of CNN such as Faster RCNN \citep{ren2015faster},
R-FCN \citep{dai2016r}, SSD \citep{liu2016ssd} and YOLO \citep{redmon2016you}.
The backbones are some feature extractors such as VGG 16 \citep{simonyan2014very}
and Resnet 101 \citep{he2016deep}. However, the number of categories
being considered usually is small: 20 for PASCAL VOC \citep{Everingham10}
and 80 for COCO \citep{lin2014microsoft}. However, those methods
are usually performed on each proposal individually without reasoning.

\textbf{Visual Reasoning. }Visual reasoning seeks to incorporate different
information or interplay between objects or scenes. Several aspects
such as shared attributes \citep{farhadi2009describing,lampert2009learning,parikh2011relative,akata2013label,almazan2014word,misra2017red},
relationships among objects can be considered. \citep{frome2013devise,mao2015learning,reed2016learning}
relies on finding similarity as the attributes in the linguistic space.
For incorporating information such as relationship, most early works
use object relations as a post-processing step \citep{torralba2003context,galleguillos2008object,felzenszwalb2010object,mottaghi2014role}.
Recent works consider a graph structure \citep{li2015gated,niepert2016learning,kipf2016semi,marino2016more,dai2017detecting,chen2018iterative}.
On the other hand, there are some sequential reasoning model for relationships
\citep{chen2017spatial,li2017attentive,chen2018iterative}. In these
works, a fixed graph is usually considered, while our module's graph
has adaptive region-to-region edges which can be embedded with any
kinds of external knowledge.

\textbf{Few-shot Recognition. }Few-shot recognition seeks to learn
a new concept with a few annotated examples which share the similar
problem with us. Early work focus on learning attributes embedding
to represent categories \citep{akata2013label,jayaraman2014zero,lampert2009learning,rohrbach2013transfer}.
Most recent works use knowledge graph such as WordNet \citep{miller1995wordnet}
to distill information among categories \citep{salakhutdinov2011learning,deng2014large,wu2016ask,marino2016more,wang2018zero}.
\citep{garcia2018few} further defined a GNN architecture to learn
a knowledge graph implicitly. In contrast, our module is explicitly
routed and benefits from the guidance of hybrid knowledge forms.

\section{The Proposed Approach}

\subsection{Overview}

The goal of this paper is to develop general modules for incorporating
knowledge to facilitate large-scale object detection with global reasoning.
Our HKRM includes two kinds of modules to support any prior knowledge
forms, shown in Figure \ref{fig:Overview-of-our_HKRM}: an explicit
knowledge module to incorporate external knowledge and an implicit
knowledge module to learn knowledge without explicit definitions or
being summarized by the human. Taking an image as the input, visual
features are extracted for each proposal region through the region
proposal network. Based on the region features, each module builds
an adaptive region-to-region undirected graph $\boldsymbol{\hat{G}}$
: $\boldsymbol{\hat{G}}=<\mathcal{N},\mathcal{\hat{\mathcal{E}}}>$,
where $\mathcal{N}$ are region proposal nodes and each edge $e_{i,j}\in\mathcal{E}$
defines a kind of knowledge between two nodes. Each module then outputs
enhanced features integrating a particular knowledge. Finally, outputs
from several modules are concatenated together and fed into the boundingbox
regression layer and classification layer to obtain final detection
results.

\subsection{Explicit Knowledge Module\label{subsec:Exhibitory-Knowledge-Module}}

\begin{figure}[t]
\begin{centering}
\includegraphics[scale=0.33]{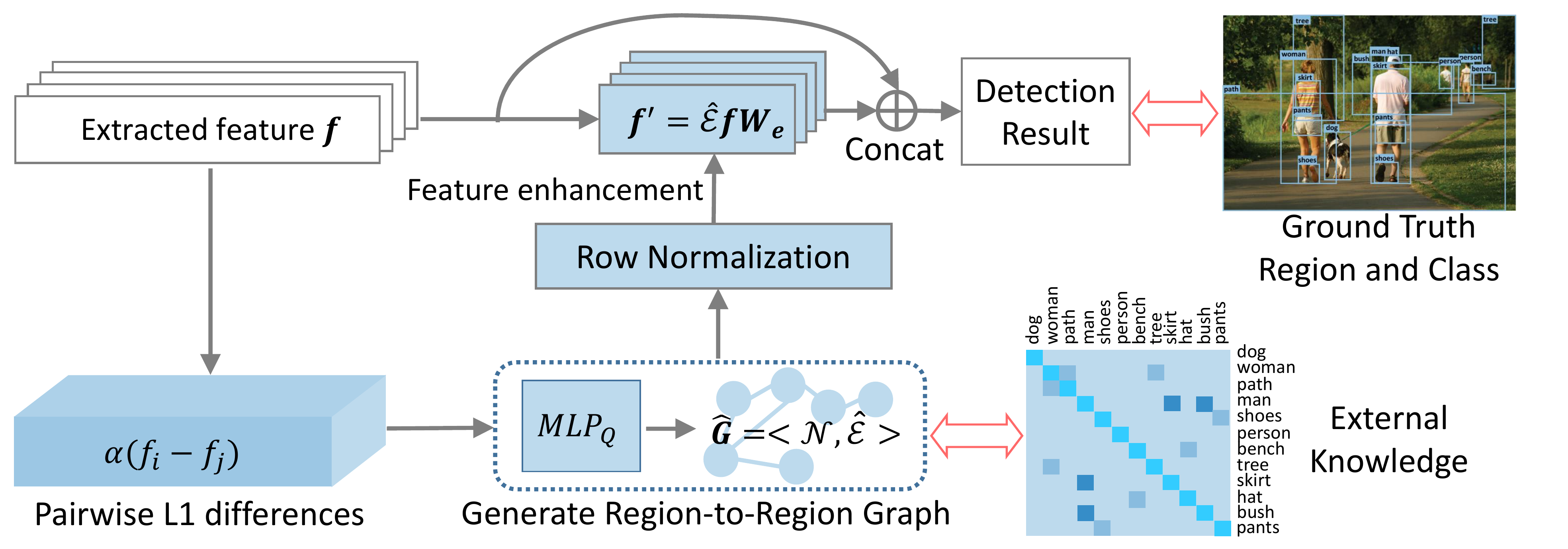}
\par\end{centering}
\caption{\label{fig:attribute}Explicit Knowledge Module. Taking the pairwise
L1 differences of the $\mathbf{f}$ as inputs, a region-to-region
graph is generated by stacked MLP. This process is supervised by the
ground truth of the external knowledge. The output evolved feature
$\mathbf{f}'$ is the enhanced feature via graph propagation. Then
$\mathbf{f}'$ is concatenated to the $\mathbf{f}$ to produce final
detection results.}

\vspace{-1mm}
\end{figure}

We regard the human commonsense knowledge that can be clearly defined
and summarized using linguistics as \emph{explicit knowledge}. The
most representative explicit knowledge forms can be attribute knowledge
(e.g. ``apple is red.'') and pairwise relationship knowledge (e.g.
``man rides bicycles''). Our explicit knowledge module aims to enhance
region features with kinds of explicit knowledge forms. Specifically,
as shown in Figure \ref{fig:attribute}, this module updates edge
connections between each pair of region graph nodes in $\hat{\boldsymbol{G}}$,
supervised by a mapping of the ground truth from a class-to-class
knowledge graph $\boldsymbol{Q}$. This $\boldsymbol{Q}$ is a certain
form of linguistic knowledge.

\subsubsection{Module Definition}

\textbf{Adaptive region-to-region graph.} We first define a region-to-region
graph $\boldsymbol{\hat{G}}$ for all $N_{r}=|\mathrm{\mathcal{N}}|$
region proposals with visual features $\mathbf{f}=\{f_{i}\}_{i=1}^{N_{r}},f_{i}\in\mathbf{\mathbb{R}}^{D}$
of $D$ dimension extracted from the backbone network, where $\mathcal{N}$
are region proposal nodes and $e_{i,j}\in\hat{\mathcal{E}}$ is the
learned graph edge for each pair of region nodes. Given any external
knowledge form, distinct edge connections $\hat{\mathcal{E}}$ can
be accordingly updated to characterize specific context information
for each region proposal in the context of specific knowledge. Formally,
given a specific knowledge graph $\boldsymbol{Q}$, each edge between
two regions $\hat{e}_{ij}$ is learned by a stacked Multi-layer Perceptron
(MLP) : 
\begin{equation}
\hat{e}_{ij}=\mathrm{MLP_{\boldsymbol{Q}}}(\alpha(f_{i},f_{j})),\label{eq:edge}
\end{equation}

where $\alpha(\cdot)$ is chosen to be the pairwise L1 difference
between features of each region pair $(f_{i},f_{j})$ since L1 difference
is symmetric. Given different prior graphs $\boldsymbol{Q}$, $\mathrm{MLP}_{\boldsymbol{Q}}$
would be parametrized with $W_{\boldsymbol{Q}}$ distinctly to generate
different region-to-region graphs $\boldsymbol{\hat{G}}$, leading
to personalized knowledge reasoning.

We learn $\mathrm{MLP}_{\boldsymbol{Q}}$ by directly enforcing the
predicted $\hat{e}_{ij}$ to be consistent with the edge weights of
a prior graph $\boldsymbol{Q}$. We define $\boldsymbol{Q}=<\mathcal{C},\mathcal{V}>$
as a class-to-class graph with $C$ class graph nodes and their prior
edge weights $v_{i,j}\mathcal{\in V}$, such as attribute and relationship
graphs. During training, as we know ground-truth categories of each
region, the edge $\hat{e}_{ij}$ of two region nodes is learned towards
the edge weights $\tilde{e}_{ij}$ of ground truth categories of region
nodes in $\boldsymbol{Q}$, that is, $\tilde{e}_{ij}=v_{c_{i},c_{j}}$
where $c_{i}$ is the ground truth class of $i$-th region. Such explicit
supervision with ground truth classes of region nodes would ensure
the learning of a reliable graph reasoning regardless of the errors
from proposal localization. $\mathrm{MLP}_{\boldsymbol{Q}}$ is then
learned to encode explicit region-wise knowledge correlations that
can be applied in the testing phase. The loss function of learned
edge weights $\{\hat{e}_{ij}\}$ for all $N_{r}$ region proposals
is defined as: 

\vspace{-2mm}
\begin{equation}
\mathcal{L}(\mathbf{f},W_{\boldsymbol{Q}},\boldsymbol{Q})=\sum_{i=1}^{N_{r}}\sum_{j=1}^{N_{r}}\frac{1}{2}(\hat{e}_{ij}-\tilde{e}_{ij})^{2}.
\end{equation}

\vspace{-2mm}

\textbf{Feature evolving via graph reasoning.} After performing row
normalization over learned edges $\hat{\mathcal{E}}=\{\hat{e}_{ij}\}$,
we can propagate features of connected regions into enhancing each
region features $\mathbf{f}'$ by different weighted edges, which
can be solved by matrix multiplication: 

\vspace{-2mm}
\begin{equation}
\mathbf{f'}=\hat{\mathcal{E}}\mathbf{f}\boldsymbol{W}_{e},
\end{equation}

\vspace{-2mm}

where $\boldsymbol{W}_{e}\in\mathbf{\mathbb{R}}^{D\times E}$ is a
transformation weight matrix and $\mathbf{f}'\in\mathbf{\mathbb{R}}^{E}$
are the enhanced features with $E$ dimension via graph reasoning.
Those regions with heavy occlusions, class ambiguities and the tiny-size
problem can be remedied by discovering adaptive contexts from other
regions guided by external knowledge. The trainable parameters are
$W_{\boldsymbol{Q}}$ of the stacked MLP and $\boldsymbol{W}_{e}$.

\subsubsection{Module Specification with Different Knowledge\label{subsec:Instantiate-the-module}}

We can specify different prior knowledge graphs $\boldsymbol{Q}$
to obtain distinct graph reasoning behaviors. Here, we take attribute
knowledge graph and relationship knowledge graph as the examples.
We refer readers to find illustrations of constructing knowledge graphs
in Supplementary material.

\textbf{Attribute Knowledge.} Attribute knowledge graph $\boldsymbol{Q^{A}}$
as one kind of $\boldsymbol{Q}$ denotes object classes are connected
with kinds of attributes such as colors, size, materials, and status.
The explicit knowledge module instantiated with attribute knowledge
will facilitate features of rare classes with more frequent classes
by transferring their shared visual attribute properties. Let us consider
$C$ classes and $K$ attributes, we obtain a $C\times K$ frequency
distribution table for each class-attribute pair, detailed in experiments.
Then the pairwise Jensen--Shannon (JS) divergence between probability
distributions $P_{c_{i}}$ and $P_{c_{j}}$ of two classes $c_{i}$
and $c_{j}$ can be measured as the edge weights of two classes $e_{c_{i},c_{j}}^{A}=JS(P_{c_{i}}||P_{c_{j}})$.
We consider JS divergence to measure the similarity instead of KL
divergence here since we expect a symmetry undirected graph while
$KL(P_{i}||P_{j})\neq KL(P_{j}||P_{i})$. Finally, the module outputs
a enhanced feature $\mathbf{f}'_{a}\in\mathbf{\mathbb{R}}^{E_{a}}$.

\textbf{Relationship Knowledge.} Relationship knowledge $\boldsymbol{Q^{R}}$
denotes the pairwise relationship between classes, such as location
relationship (e.g. \emph{along}, \emph{on}), the ``subject-verb-object''
relationship (e.g. \emph{eat}, \emph{wear}) or co-occurrence relationship.
The evolved features will be enhanced with high-level semantic correlations
between regions. Similarly, we obtain $\boldsymbol{Q}^{\boldsymbol{R}}$
by calculating frequent statistics either from the semantic information
or simply from the occurrence among all class pairs. The symmetric
transformation and row normalization are performed on edge weights.
Let $\mathbf{f}'_{r}\in\mathbf{\mathbb{R}}^{E_{r}}$ denotes the output
of the explicit relationship module.

\subsection{Implicit Knowledge Module}

Considering some commonsense knowledge without explicit definitions
or being summarized by the human, we regard them as \emph{implicit
knowledge} and thus an implicit knowledge module is designed. Taking
geometry priors as an example, besides those explicit pairwise locations,
there also exists some complicate location information, such as ``the
ceiling is always above all the other objects'' and “the water is
always below the ships, mountains and the sky”. Taking features $\mathbf{q}=\{q_{i}\}$
as inputs that depict the features of each region (e.g. geometric
features), our implicit knowledge module integrates multiple graph
reasoning over $M$ region-to-region graphs obtained by M stacked
MLPs following (1) to encode these implicit priors. The analogous
idea of multi-head attention can be found in \citep{chen2018iterative,hu2017relation,Vaswani2017Attention}.
This enables the module to catch multiple spatial relationships such
as ``up and down'', ``left and right'' and ``corner and center''.
Visualization of different learned graphs can be found in Supplementary
material. Similar to region-to-region graph used in explicit knowledge
module, we learn specific edge weights $\{\hat{e}_{ij}^{(m)}\}$ of
each graph $\hat{\mathbf{G}}_{m},m=1,\dots,M$ for all-region proposal
pairs, following Eqn. \ref{eq:edge}. We then average edge weights
of all graph $\{\hat{\mathbf{G}}_{m}\}$ and add them with a identity
matrix $\boldsymbol{I}$ to obtain the edge connections $\hat{e}_{ij}^{I}\in\hat{\mathcal{E}}^{I}$: 

\vspace{-2mm}
\begin{equation}
\hat{e}_{ij}^{I}=\frac{1}{M}\sum_{m=1}^{M}\hat{e}_{ij}^{(m)}+\boldsymbol{I}.
\end{equation}

\vspace{-2mm}
We then adopt matrix multiplication $\mathbf{g}'=\hat{\mathcal{E}}^{I}\mathbf{f}\boldsymbol{W}_{g}$
to get the evolved features $\mathbf{g}'\in\mathbf{\mathbb{R}}^{E_{g}}$.
The trainable parameters are weights of $M$ stacked MLP for learning
edge weights of knowledge graphs $\{\hat{\mathbf{G}}_{m}\}$, and
the transformation matrix $\boldsymbol{W}_{g}\in\mathbf{\mathbb{R}}^{D\times E_{g}}$
is shared for all graphs.

\textbf{Module specification with spatial layout.} Here, we instantiate
the implicit knowledge module by spatial layout inputs to capture
complicated spatial constraints by using specific input information.
The input geometry feature $q_{i}$ of each region is simply object
bounding box. To make $q_{i}$ be invariant to the scale transformation,
a relative geometry feature is used, as $(\frac{x_{i}}{\bar{w}},\frac{y_{i}}{\bar{h}},\frac{w_{i}}{\bar{w}},\frac{h_{i}}{\bar{h}},p_{i})$,
where $\bar{w}$ and $\bar{h}$ denotes the size of the image and
$p_{i}$ is the initial foreground probability of each region. Note
that edge weights are implicitly learned via the back-propagation
of the whole network.

\section{Experiments}

\begin{table}
\begin{centering}
\tabcolsep 0.01in{\scriptsize{}}%
\begin{tabular}{l|l|llllll|llllll}
\hline 
{\scriptsize{}\%} & {\scriptsize{}Method} & {\scriptsize{}AP} & {\scriptsize{}AP$_{50}$} & {\scriptsize{}AP$_{75}$} & {\scriptsize{}AP$_{S}$} & {\scriptsize{}AP$_{M}$} & {\scriptsize{}AP$_{L}$} & {\scriptsize{}AR$_{1}$} & {\scriptsize{}AR$_{10}$} & {\scriptsize{}AR$_{100}$} & {\scriptsize{}AR$_{S}$} & {\scriptsize{}AR$_{M}$} & {\scriptsize{}AR$_{L}$}\tabularnewline
\hline 
\multirow{7}{*}{{\scriptsize{}\begin{sideways}{\scriptsize{}VG$_{1000}$}\end{sideways}}} & {\scriptsize{}Light-head rcnn\citep{li2017light}} & {\scriptsize{}6.2} & {\scriptsize{}10.9} & {\scriptsize{}6.2} & {\scriptsize{}2.8} & {\scriptsize{}6.5} & {\scriptsize{}9.8} & {\scriptsize{}14.6} & {\scriptsize{}18.0} & {\scriptsize{}18.7} & {\scriptsize{}7.2} & {\scriptsize{}17.1} & {\scriptsize{}25.3}\tabularnewline
 & {\scriptsize{}FPN\citep{lin2017feature}} & {\scriptsize{}5.6} & {\scriptsize{}10.1} & {\scriptsize{}5.4} & {\scriptsize{}3.4} & {\scriptsize{}5.8} & {\scriptsize{}8.0} & {\scriptsize{}13.0} & {\scriptsize{}16.5} & {\scriptsize{}16.6} & {\scriptsize{}7.5} & {\scriptsize{}15.7} & {\scriptsize{}20.6}\tabularnewline
\cline{2-14} 
 & {\scriptsize{}Faster RCNN\citep{ren2015faster}} & {\scriptsize{}5.8} & {\scriptsize{}10.7} & {\scriptsize{}5.7} & {\scriptsize{}1.9} & {\scriptsize{}5.8} & {\scriptsize{}10.0} & {\scriptsize{}13.7} & {\scriptsize{}17.2} & {\scriptsize{}17.2} & {\scriptsize{}4.9} & {\scriptsize{}15.7} & {\scriptsize{}25.3}\tabularnewline
 & {\scriptsize{}Attribute} & {\scriptsize{}7.4} & {\scriptsize{}12.9} & {\scriptsize{}7.4} & {\scriptsize{}2.4} & {\scriptsize{}7.4} & {\scriptsize{}13.7} & {\scriptsize{}17.0} & {\scriptsize{}21.4} & {\scriptsize{}21.5} & {\scriptsize{}6.0} & {\scriptsize{}19.5} & {\scriptsize{}33.0}\tabularnewline
 & {\scriptsize{}Relation} & {\scriptsize{}7.4} & {\scriptsize{}12.8} & {\scriptsize{}7.5} & {\scriptsize{}3.0} & {\scriptsize{}7.5} & {\scriptsize{}13.0} & {\scriptsize{}17.0} & {\scriptsize{}21.6} & {\scriptsize{}21.7} & {\scriptsize{}7.2} & {\scriptsize{}19.8} & {\scriptsize{}31.4}\tabularnewline
 & {\scriptsize{}Spatial} & {\scriptsize{}7.3} & {\scriptsize{}12.1} & {\scriptsize{}7.7} & {\scriptsize{}2.7} & {\scriptsize{}7.2} & {\scriptsize{}12.7} & {\scriptsize{}17.7} & {\scriptsize{}21.9} & {\scriptsize{}22.0} & {\scriptsize{}6.3} & {\scriptsize{}19.5} & \textbf{\scriptsize{}33.3}\tabularnewline
 & {\scriptsize{}HKRM (All)} & \textbf{\scriptsize{}7.8}{\scriptsize{}$^{+2.0}$} & \textbf{\scriptsize{}13.4}{\scriptsize{}$^{+2.7}$} & \textbf{\scriptsize{}8.1}{\scriptsize{}$^{+2.4}$} & \textbf{\scriptsize{}4.1}{\scriptsize{}$^{+2.2}$} & \textbf{\scriptsize{}8.1}{\scriptsize{}$^{+2.3}$} & \textbf{\scriptsize{}12.7}{\scriptsize{}$^{+2.7}$} & \textbf{\scriptsize{}18.1}{\scriptsize{}$^{+4.4}$} & \textbf{\scriptsize{}22.7}{\scriptsize{}$^{+5.5}$} & \textbf{\scriptsize{}22.7}{\scriptsize{}$^{+5.5}$} & \textbf{\scriptsize{}9.6}{\scriptsize{}$^{+4.7}$} & \textbf{\scriptsize{}20.8}{\scriptsize{}$^{+5.1}$} & {\scriptsize{}31.4$^{+6.1}$}\tabularnewline
\hline 
\multirow{7}{*}{{\scriptsize{}\begin{sideways}{\scriptsize{}VG$_{3000}$}\end{sideways}}} & {\scriptsize{}Light-head rcnn\citep{li2017light}} & {\scriptsize{}3.0} & {\scriptsize{}5.1} & {\scriptsize{}3.2} & {\scriptsize{}1.7} & {\scriptsize{}4.0} & {\scriptsize{}5.8} & {\scriptsize{}7.3} & {\scriptsize{}9.0} & {\scriptsize{}9.0} & {\scriptsize{}4.3} & {\scriptsize{}10.3} & {\scriptsize{}15.4}\tabularnewline
 & {\scriptsize{}FPN\citep{lin2017feature}} & {\scriptsize{}3.3} & {\scriptsize{}5.2} & {\scriptsize{}3.2} & {\scriptsize{}1.9} & {\scriptsize{}4.3} & {\scriptsize{}4.8} & {\scriptsize{}6.9} & {\scriptsize{}8.3} & {\scriptsize{}8.3} & {\scriptsize{}4.3} & {\scriptsize{}9.8} & {\scriptsize{}11.6}\tabularnewline
\cline{2-14} 
 & {\scriptsize{}Faster RCNN\citep{ren2015faster}} & {\scriptsize{}3.4} & {\scriptsize{}6.0} & {\scriptsize{}3.4} & {\scriptsize{}1.6} & {\scriptsize{}4.3} & {\scriptsize{}7.3} & {\scriptsize{}8.1} & {\scriptsize{}9.8} & {\scriptsize{}9.8} & {\scriptsize{}3.8} & {\scriptsize{}10.9} & {\scriptsize{}17.0}\tabularnewline
 & {\scriptsize{}Attribute} & {\scriptsize{}4.1} & {\scriptsize{}7.0} & {\scriptsize{}4.3} & {\scriptsize{}2.5} & {\scriptsize{}5.3} & {\scriptsize{}7.9} & {\scriptsize{}9.7} & {\scriptsize{}11.7} & {\scriptsize{}11.7} & {\scriptsize{}5.7} & {\scriptsize{}12.8} & {\scriptsize{}19.6}\tabularnewline
 & {\scriptsize{}Relation} & {\scriptsize{}4.2} & {\scriptsize{}7.1} & {\scriptsize{}4.3} & {\scriptsize{}2.6} & {\scriptsize{}5.3} & {\scriptsize{}8.1} & {\scriptsize{}9.7} & {\scriptsize{}11.9} & {\scriptsize{}11.9} & {\scriptsize{}6.0} & {\scriptsize{}12.8} & {\scriptsize{}19.8}\tabularnewline
 & {\scriptsize{}Spatial} & {\scriptsize{}4.0} & {\scriptsize{}6.7} & {\scriptsize{}4.1} & {\scriptsize{}2.3} & {\scriptsize{}5.1} & {\scriptsize{}7.6} & {\scriptsize{}9.3} & {\scriptsize{}11.2} & {\scriptsize{}11.2} & {\scriptsize{}5.3} & {\scriptsize{}12.4} & {\scriptsize{}18.7}\tabularnewline
 & {\scriptsize{}HKRM (All)} & \textbf{\scriptsize{}4.3$^{+0.9}$} & \textbf{\scriptsize{}7.2$^{+1.2}$} & \textbf{\scriptsize{}4.4$^{+1.0}$} & \textbf{\scriptsize{}2.6$^{+1.0}$} & \textbf{\scriptsize{}5.5$^{+1.2}$} & \textbf{\scriptsize{}8.4$^{+1.1}$} & \textbf{\scriptsize{}10.1$^{+2.0}$} & \textbf{\scriptsize{}12.2$^{+2.4}$} & \textbf{\scriptsize{}12.2$^{+2.4}$} & \textbf{\scriptsize{}5.9$^{+2.1}$} & \textbf{\scriptsize{}13.0$^{+2.1}$} & \textbf{\scriptsize{}20.5$^{+2.5}$}\tabularnewline
\hline 
\multirow{7}{*}{{\scriptsize{}\begin{sideways}{\scriptsize{}ADE}\end{sideways}}} & {\scriptsize{}Light-head rcnn\citep{li2017light}} & {\scriptsize{}7.0} & {\scriptsize{}11.7} & {\scriptsize{}7.3} & {\scriptsize{}2.4} & {\scriptsize{}5.1} & {\scriptsize{}11.2} & {\scriptsize{}9.6} & {\scriptsize{}13.3} & {\scriptsize{}13.4} & {\scriptsize{}4.3} & {\scriptsize{}10.4} & {\scriptsize{}20.4}\tabularnewline
 & {\scriptsize{}FPN\citep{lin2017feature}} & {\scriptsize{}6.5} & {\scriptsize{}12.1} & {\scriptsize{}6.2} & {\scriptsize{}3.3} & {\scriptsize{}6.0} & {\scriptsize{}10.5} & {\scriptsize{}9.5} & {\scriptsize{}12.9} & {\scriptsize{}13.0} & {\scriptsize{}5.3} & {\scriptsize{}11.9} & {\scriptsize{}18.6}\tabularnewline
\cline{2-14} 
 & {\scriptsize{}Faster RCNN\citep{ren2015faster}} & {\scriptsize{}7.9} & {\scriptsize{}14.7} & {\scriptsize{}7.5} & {\scriptsize{}2.1} & {\scriptsize{}5.8} & {\scriptsize{}13.2} & {\scriptsize{}10.6} & {\scriptsize{}14.2} & {\scriptsize{}14.4} & {\scriptsize{}4.5} & {\scriptsize{}11.9} & {\scriptsize{}22.4}\tabularnewline
 & {\scriptsize{}Attribute} & {\scriptsize{}9.6} & {\scriptsize{}16.8} & {\scriptsize{}9.7} & {\scriptsize{}3.1} & {\scriptsize{}7.0} & {\scriptsize{}15.9} & {\scriptsize{}12.7} & {\scriptsize{}16.9} & {\scriptsize{}17.1} & {\scriptsize{}6.1} & {\scriptsize{}14.1} & {\scriptsize{}26.3}\tabularnewline
 & {\scriptsize{}Relation} & {\scriptsize{}9.6} & {\scriptsize{}16.8} & {\scriptsize{}9.8} & {\scriptsize{}3.0} & {\scriptsize{}7.2} & {\scriptsize{}15.4} & {\scriptsize{}12.6} & {\scriptsize{}16.8} & {\scriptsize{}17.0} & {\scriptsize{}6.2} & {\scriptsize{}14.2} & {\scriptsize{}26.0}\tabularnewline
 & {\scriptsize{}Spatial} & {\scriptsize{}8.7} & {\scriptsize{}14.0} & {\scriptsize{}9.0} & {\scriptsize{}3.1} & {\scriptsize{}6.9} & {\scriptsize{}14.3} & {\scriptsize{}11.4} & {\scriptsize{}15.5} & {\scriptsize{}15.6} & {\scriptsize{}5.0} & {\scriptsize{}12.7} & {\scriptsize{}24.2}\tabularnewline
 & {\scriptsize{}HKRM (All)} & \textbf{\scriptsize{}10.3}{\scriptsize{}$^{+2.4}$} & \textbf{\scriptsize{}18.0}{\scriptsize{}$^{+3.0}$} & \textbf{\scriptsize{}10.4}{\scriptsize{}$^{+2.9}$} & \textbf{\scriptsize{}4.1}{\scriptsize{}$^{+2.0}$} & \textbf{\scriptsize{}7.9}{\scriptsize{}$^{+2.1}$} & \textbf{\scriptsize{}16.8}{\scriptsize{}$^{+3.6}$} & \textbf{\scriptsize{}13.6}{\scriptsize{}$^{+3.0}$} & \textbf{\scriptsize{}18.3}{\scriptsize{}$^{+4.1}$} & \textbf{\scriptsize{}18.5}{\scriptsize{}$^{+4.1}$} & \textbf{\scriptsize{}7.1}{\scriptsize{}$^{+2.6}$} & \textbf{\scriptsize{}15.5}{\scriptsize{}$^{+3.6}$} & \textbf{\scriptsize{}28.4}{\scriptsize{}$^{+6.0}$}\tabularnewline
\hline 
\end{tabular}{\scriptsize\par}
\par\end{centering}
\vspace{1mm}

\caption{\label{tab:Main-results-onVG_ADE}Main results of test datasets on
VG$_{1000}$ , VG$_{3000}$ and ADE. ``Attribute'', Relation''
and ``Spatial'' are the baseline Faster RCNN adding the corresponding
knowledge module alone. HKRM is the model with a combination of all.}

\vspace{-4mm}
\end{table}
\textbf{Dataset and Evaluation.} We conduct experiments on large-scale
object detection benchmarks with a large number of classes: that is,
Visual Genome (VG) \citep{krishnavisualgenome} and ADE \citep{zhou2017scene}.
The task is to localize an object and classify it, which is different
from the experiments with given ground truth locations \citep{chen2018iterative}.
For Visual Genome, we use the latest release (v1.4), and synsets \citep{russakovsky2015imagenet}
instead of the raw names of the categories due to inconsistent label
annotations, following \citep{hu2018learning,chen2018iterative}.
We consider two set of target classes: 1000 most frequent classes
and 3000 most frequent classes, resulting in two settings $\text{VG}_{1000}$
and $\text{VG}_{3000}$. We split the remaining 92960 images with
objects on these class sets into 87960 and 5,000 for training and
testing, respectively. In term of ADE dataset, we use 20,197 images
for training and 1,000 images for testing, following \citep{chen2018iterative}.
To validate the generalization capability of models and the usefulness
of transferred knowledge graph from VG, $445$ classes that overlap
with VG dataset are selected as targets. Since ADE is a segmentation
dataset, we convert segmentation masks to bounding boxes \citep{chen2018iterative}
for all instances. For evaluation, we adopt the metrics from COCO
detection evaluation criteria \citep{lin2014microsoft}, that is,
mean Average Precision (mAP) across different IoU thresholds (IoU$=\{0.5:0.95,0.5,0.75\}$)
and scales (small, medium, big). We also use Average Recall (AR) with
different number of given detection per image ($\left\{ 1,10,100\right\} $)
and different scales (small, medium, big). 

Additionally, we also evaluate on PASCAL VOC 2007 \citep{Everingham10}
and MSCOCO 2017 \citep{lin2014microsoft} to show prior knowledge
can help detection for a small set of frequent classes (20/80 classes).
PASCAL VOC consists of about 10k trainval images (included VOC 2007
trainval and VOC 2012 trainval) and 5k test images over 20 object
categories. We only report mAP scores using IoU thresholds at $0.5$
for the purpose of comparison with other existing methods. MSCOCO
2017 contains 118k images for training, 5k for evaluation.

\textbf{Knowledge Graph Construction.} We apply general knowledge
graphs for both experiments on VG and ADE datasets. With the help
of the statistics of the annotations in the VG dataset, we can both
create attribute knowledge and relationship knowledge graphs. Specifically,
we consider top 200 most frequent attributes annotations in VG such
as color, material and status of the categories ($C=3000$), and then
count their frequent statistics as the class-attribute table. For
relationship knowledge, we use top 200 most frequent relationship
annotations in VG such as location relationship, subject-verb-object
relationship, and count frequent statistics of each class-relationship
pair. Illustrations of constructed knowledge graphs can be found in
Supplementary material.

\textbf{Implementation Details.} We treat the state-of-the-art Faster
RCNN \citep{ren2015faster,jjfaster2rcnn} as our baseline and implement
all models in Pytorch \citep{paszke2017automatic}. We also compare
with Light-head RCNN \citep{li2017light} and FPN \citep{lin2017feature}.
ResNet-101 \citep{he2016deep} pretrained on ImageNet \citep{russakovsky2015imagenet}
is used as our backbone network. The parameters before conv3 and the
batch normalization are fixed, same with \citep{li2017light}. During
training, we augment with flipped images and multi-scaling (pixel
size=$\left\{ 400,500,600,700,800\right\} $). During testing, pixel
size$=600$ is used. Following \citep{ren2015faster}, RPN is applied
on the conv4 feature maps. The total number of proposed regions after
NMS is $128$. Features in conv5 are avg-pooled to become the input
of the final classifier. Unless otherwise noted, settings are same
for all experiments. In terms of our explicit attribute and relationship
knowledge module upon region proposals, we use the final conv5 for
$128$ regions after avg-pool ($D$$=2048$) as our module inputs.
We consider a $4$ stacked linear layers as $\mathrm{MLP_{\boldsymbol{Q}}}$(output
channels:$\left[256,128,64,1\right]$). ReLU is selected as the activation
function between each linear layer. The output size : $E_{a}=E_{r}=256$,
which is considered sufficient to contain the enhanced feature. In
terms of implicit knowledge module, we employ $M=10$ implicit graphs.
For learning each graph, 2 stacked linear layers are used (output
channels:$\left[5,1\right]$). $p_{i}$ is the score of the foreground
form the RPN. The output size: $E_{g}=256$. To avoid over-fitting,
the final version of HKRM is the combination of three shrink modules
with each output size equals $256$. $\mathbf{f}'_{a}$, $\mathbf{f}'_{r}$,
$\mathbf{g}'$ and $\mathbf{f}$ are concatenated together and fed
into the boundingbox regression layer and classification layer.We
apply stochastic gradient descent with momentum to optimize all models.
The initial learning rate is 0.01, reduce three times ($\times0.01$)
during fine-tuning; $10^{-4}$ as weight decay; 0.9 as momentum. For
both VG and ADE dataset, we train 28 epochs with mini-batch size of
2 for both the baseline Faster RCNN. (Further training after 14 epochs
won't increase the performance of baseline.) For our HKRM, we use
14 epochs of the baseline as pretrained model and train another 14
epochs with same settings with baseline.

\begin{table}
\begin{centering}
\tabcolsep 0.01in{\small{}}%
\begin{tabular}{c|c|c|c|c}
\hline 
{\small{}Dataset} & {\small{}Method} & {\small{}Backbone} & {\small{}\#. Parameter (M) } & {\small{}mAP (\%)}\tabularnewline
\hline 
\multirow{3}{*}{{\small{}PASCAL VOC$_{20}$}} & {\small{}SMN\citep{chen2017spatial}} & {\small{}ResNet-101} & {\small{}66.7} & {\small{}67.8}\tabularnewline
 & {\small{}Faster RCNN\citep{ren2015faster}} & {\small{}ResNet-101} & {\small{}57.0} & {\small{}75.1}\tabularnewline
 & {\small{}HKRM (All)} & {\small{}ResNet-101} & {\small{}59.2} & \textbf{\small{}78.8}\tabularnewline
\hline 
\multirow{4}{*}{{\small{}MSCOCO$_{80}$}} & {\small{}SMN\citep{chen2017spatial}} & {\small{}ResNet-101} & {\small{}68.1} & {\small{}31.6}\tabularnewline
 & {\small{}Relation Network\citep{hu2017relation}} & {\small{}ResNet-101} & {\small{}64.6} & {\small{}35.2}\tabularnewline
 & {\small{}Faster RCNN\citep{ren2015faster}} & {\small{}ResNet-101} & {\small{}58.3} & {\small{}34.2}\tabularnewline
 & {\small{}HKRM (All)} & {\small{}ResNet-101} & {\small{}60.3} & \textbf{\small{}37.8}\tabularnewline
\hline 
\end{tabular}{\small\par}
\par\end{centering}
\vspace{1mm}

\caption{\label{coco_vov_parameter}Comparisons of mean Average Precision (mAP)
and \#. Parameter on PASCAL VOC 2007 test set and COCO 2017 val set. }

\vspace{-4mm}
\end{table}

\subsection{Comparison with state-of-the-art}

\begin{figure}[t]
\begin{centering}
\begin{sideways}{\hspace{2 mm}Faster RCNN}\end{sideways} \includegraphics[scale=0.25]{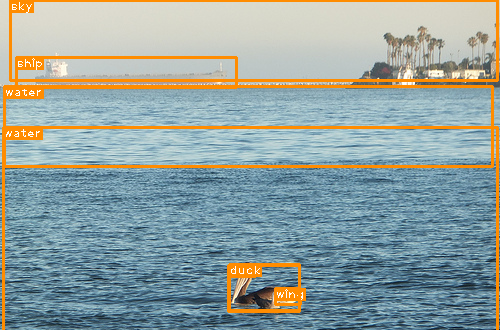}\includegraphics[scale=0.25]{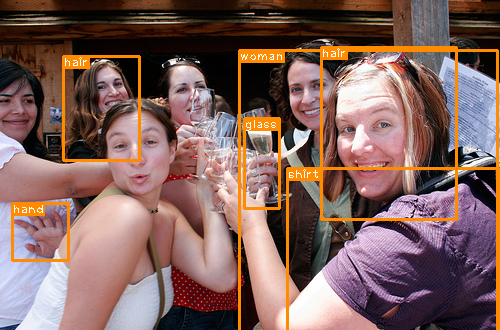}\includegraphics[scale=0.25]{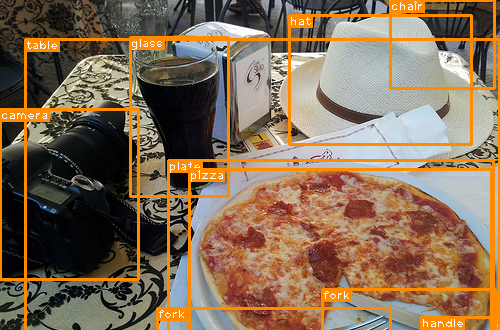}\includegraphics[scale=0.25]{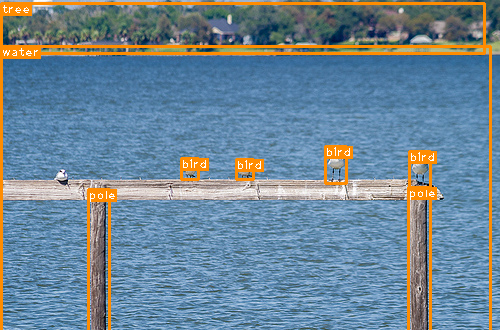}
\par\end{centering}
\begin{centering}
\begin{sideways}{\hspace{5 mm}HKRM}\end{sideways} \includegraphics[scale=0.25]{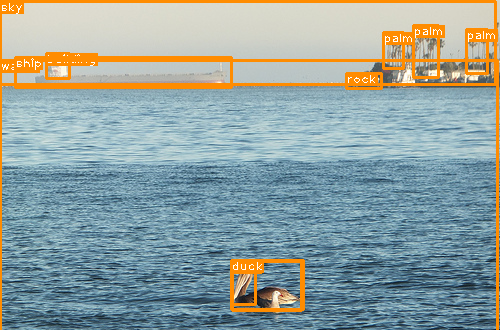}\includegraphics[scale=0.25]{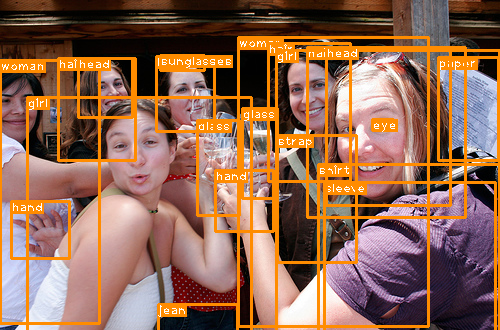}\includegraphics[scale=0.25]{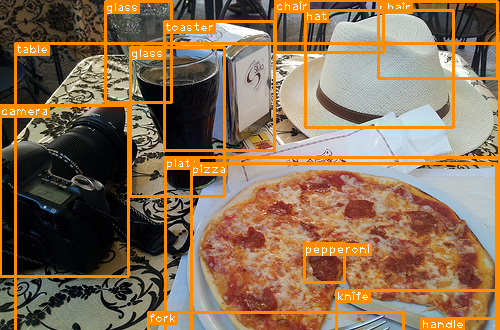}\includegraphics[scale=0.25]{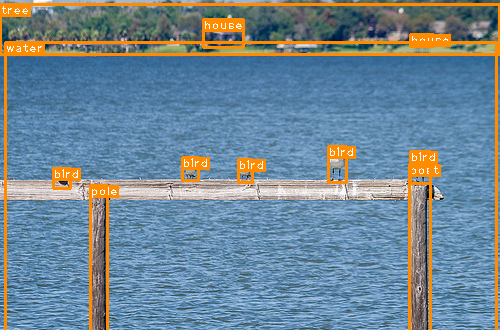}
\par\end{centering}
\caption{\label{fig:Representative-examples}Qualitative result comparison
on VG$_{1000}$ between Faster RCNN and our HKRM. Objects with occlusion,
ambiguities and rare category can be detected by our modules.}

\vspace{-3mm}
\end{figure}
We report the result comparisons on VG$_{1000}$ with 1000 categories
, VG$_{3000}$ with 3000 categories and ADE dataset in Table \ref{tab:Main-results-onVG_ADE}.
As can be seen, all our model variants outperform the baseline Faster
RCNN\citep{ren2015faster} on all dataset. Our HKRM achieves an overall
AP of 7.8\% compared to 5.8\% by Faster RCNN on VG$_{1000}$, 4.3\%
compared to 3.4\% on VG$_{3000}$, and 10.3\% compared to 7.9\% on
ADE, respectively. Moreover, our model achieves significant higher
performance on both classification and localization accuracy than
the baseline on all cases (i.e. different scales and overlaps). This
verifies the effectiveness of incorporating global reasoning guided
by rich external knowledge into local region recognition. More significant
performance gap by our HKRM can be observed for those rare categories
with very few samples (about 1.5\% average improvement for the top
150 infrequent categories by our method in terms of mAP). 

To compare with the state-of-art knowledge-enhanced methods, we also
implement HKRM on PASCAL VOC and MS COCO datasets with only 20/80
categories in Table \ref{coco_vov_parameter}. For PASCAL VOC, our
HKRM performs 1.1\% better than the baseline Faster RCNN, and outperforms
Spatial Memory Network \citep{chen2017spatial}. For MSCOCO, comparison
is made between Relation Network \citep{hu2017relation} and Spatial
Memory Network. The proposed HKRM boosts the mAP from 34.9\% to 37.8\%
and outperform all the other methods. Our method can also boost the
performance in the more simplified dataset benefiting from the shared
linguistic knowledge and spatial layout knowledge. Note that HKRM
consisted of three knowledge modules totally increases about 2\% parameters
and is light-weight compared to \citep{chen2017spatial,hu2017relation}. 

Figure~\ref{fig:Representative-examples} shows the qualitative result
comparison between our HKRM and Faster RCNN. Our HKRM can detect the
obscure palm trees far away in the left image. In the middle image,
the multiple overlapped small objects such as glass and paper is recognized
by our method. ``Pepperoni'' is a rare category and is detected
on the pizza in the right image.

\subsection{Ablation Studies}

\textbf{The effect of different explicit knowledge.} We analyze the
effect of both attribute and relationship knowledge on final detection
performance. The attribute module along can increase overall AP by
1.6\% for VG$_{1000}$, 0.6\% for VG$_{3000}$ and 1.7\% for ADE over
baseline. The relationship module has similar performance with a slightly
higher result for VG$_{3000}$. Sharing visual feature according to
both attribute and relationship knowledge can actually boost the performance
of object detection.

\textbf{The effect of different explicit knowledge.} We analyze the
effect of both attribute knowledge and relationship knowledge on final
detection performance. The attribute module along can increase overall
AP by 1.6\% for VG$_{1000}$, 0.6\% for VG$_{3000}$ and 1.7\% for
ADE over baseline. The relationship module has similar performance
with a slightly higher result for VG$_{3000}$. Sharing visual feature
according to both attribute and relationship knowledge can actually
boost the performance of object detection.

\textbf{The effect of implicit knowledge.} As can be seen, the implicit
spatial module alone helps around 1.5\% for VG$_{1000}$, 0.3\% for
VG$_{3000}$ and 0.8\% for ADE. The spatial module alone is not as
effective 263 as the attribute and relation module. However, the unsupervised
learning of the spatial knowledge 264 still can significantly help
the object recognition through those undefined knowledge.

\textbf{Generalization capability.} From Table \ref{tab:Main-results-onVG_ADE},
the external knowledge graph from VG can actually help to improve
the performance of ADE. Therefore, any datasets with overlap categories
can share the existing knowledge graph. Besides, our module can be
added to diverse detection systems easily.

\textbf{Global reasoning.} The proposed HKRM achieves the global reasoning
over regions via one-time propagation over all graph edges and nodes.
Benefiting from the learned knowledge graph for each image, our HKRM
is able to propagate information between nodes which are not connected
in the prior knowledge graph. We have tried the higher orders of feature
transformation (e.g. 2 and 3) and did not observed significant improvement.
In fact, over-transformation will even make the enhanced features
all identical.

\begin{figure}[t]
\begin{centering}
\includegraphics[scale=0.8]{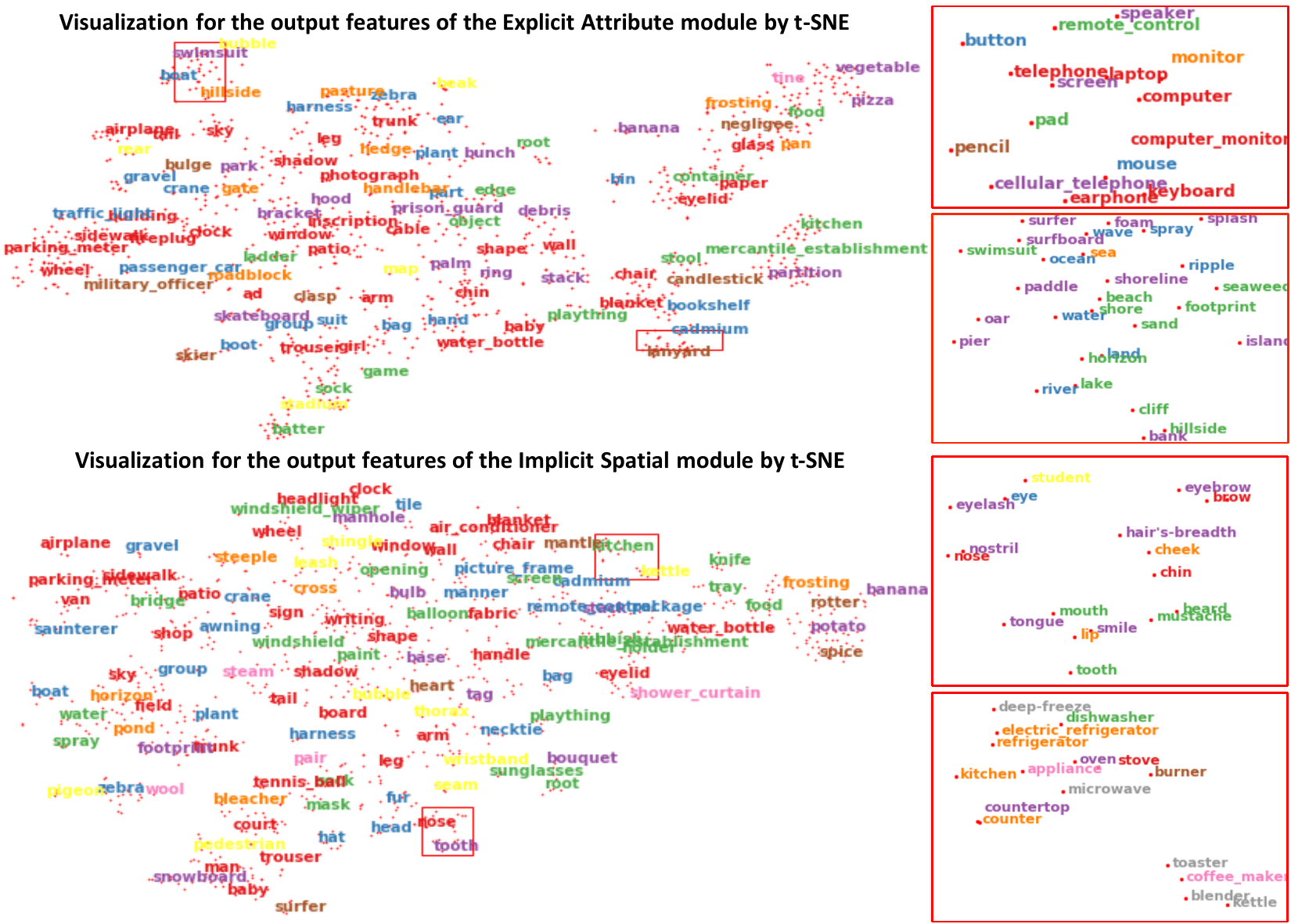}
\par\end{centering}
\vspace{-1mm}

\caption{\label{fig:Visualization-fa-fs}2-D visualization of $\mathbf{f}'_{a}$
and $\mathbf{g}'$ by t-SNE method \citep{maaten2008visualizing}:
the explicit module with attribute knowledge (top); implicit knowledge
module with spatial knowledge (bottom) . The red regions are enlarged
in right panels. The categories shared the similar attribute knowledge
(top) and spatial relationship (bottom) are closed to each other.
This verifies that our modules learn the corresponding knowledge.}

\vspace{-3mm}
\end{figure}

\textbf{Analysis of feature interpretability.} To better understand
the underlying feature representations that our HKRM actually learn
for graph reasoning, we record the output $\mathbf{f}'_{a}$ and $\mathbf{g}'$
($E_{a}=E_{g}=512$) from the explicit attribute module and implicit
spatial module and its corresponding real labels from each region
of $8000$ VG$_{1000}$ images. Then we take average according to
the labels and use the t-SNE \citep{maaten2008visualizing} clustering
method to visualize them as shown in Figure \ref{fig:Visualization-fa-fs}.
Note that if features of some categories are closed to each other,
the edges between those close categories are more likely to be activated.
From two enlarged regions on top, we can see that features of categories
which share similar attributes such as ``water'', ''sand'' and
``electronics'' are closed to each other. And this speaks well our
explicit knowledge module successfully incorporates the prior attribute
knowledge and leads to interpretable feature learning. Similarly,
from two bottom enlarged regions, features of objects which has spatial
relationship such as ``on face'' and ``in kitchen counter'' are
closed to each other. This validates our spatial knowledge module
is capable of encoding underlying spatial relationships. Benefiting
from explicit knowledge supervision, the feature clustering property
of the explicit attribute module seems to be better than those of
the implicit knowledge module. More gradient visualization \citep{springenberg2014striving}
results for the enhanced features are included in Supplementary materials
for better understanding the module.

\section{Conclusion}

We present two novel general knowledge modules in HKRM. The first
one can embed any external knowledge through supervision. The second
one can implicitly learn some knowledge without explicit definitions
or being summarized by human. Both modules can be easily applied to
the original detection system to improve the detection performance.
The experiment and analysis indicated HKRM can alleviate the problems
of large-scale object detection. For our future work, we can use Cholesky
decomposition to re-parametrize the region-to-region graph to further
reduce half of the module parameters due to the property of symmetry
of our graph. We can also add experiments using the word embedding
knowledge in the explicit module and the latest new Open Images Dataset
which consists about 600 categories.

\subsubsection*{Acknowledgments}

This work was supported in part by the National Key Research and Development Program of China under Grant No. 2018YFC0830103, in part by National High Level Talents Special Support Plan (Ten Thousand Talents Program), and in part by National Natural Science Foundation of China (NSFC) under Grant No. 61622214, and 61836012.

\bibliographystyle{abbrv}
\bibliography{Know_network}

\begin{thebibliography}{10}

\bibitem{akata2013label}
Z.~Akata, F.~Perronnin, Z.~Harchaoui, and C.~Schmid.
\newblock Label-embedding for attribute-based classification.
\newblock In {\em CVPR}, 2013.

\bibitem{almazan2014word}
J.~Almaz{\'a}n, A.~Gordo, A.~Forn{\'e}s, and E.~Valveny.
\newblock Word spotting and recognition with embedded attributes.
\newblock {\em IEEE transactions on pattern analysis and machine intelligence},
  36(12):2552--2566, 2014.

\bibitem{biederman1982scene}
I.~Biederman, R.~J. Mezzanotte, and J.~C. Rabinowitz.
\newblock Scene perception: Detecting and judging objects undergoing relational
  violations.
\newblock {\em Cognitive psychology}, 14(2):143--177, 1982.

\bibitem{cai2017cascade}
Z.~Cai and N.~Vasconcelos.
\newblock Cascade r-cnn: Delving into high quality object detection.
\newblock In {\em CVPR}, 2018.

\bibitem{chen2017spatial}
X.~Chen and A.~Gupta.
\newblock Spatial memory for context reasoning in object detection.
\newblock In {\em ICCV}, 2017.

\bibitem{chen2018iterative}
X.~Chen, L.-J. Li, L.~Fei-Fei, and A.~Gupta.
\newblock Iterative visual reasoning beyond convolutions.
\newblock In {\em CVPR}, 2018.

\bibitem{dai2017detecting}
B.~Dai, Y.~Zhang, and D.~Lin.
\newblock Detecting visual relationships with deep relational networks.
\newblock In {\em CVPR}, 2017.

\bibitem{dai2016r}
J.~Dai, Y.~Li, K.~He, and J.~Sun.
\newblock R-fcn: Object detection via region-based fully convolutional
  networks.
\newblock In {\em NIPS}, 2016.

\bibitem{deng2014large}
J.~Deng, N.~Ding, Y.~Jia, A.~Frome, K.~Murphy, S.~Bengio, Y.~Li, H.~Neven, and
  H.~Adam.
\newblock Large-scale object classification using label relation graphs.
\newblock In {\em ECCV}, 2014.

\bibitem{Everingham10}
M.~Everingham, L.~Van~Gool, C.~K.~I. Williams, J.~Winn, and A.~Zisserman.
\newblock The pascal visual object classes (voc) challenge.
\newblock {\em International Journal of Computer Vision}, 88(2):303--338, June
  2010.

\bibitem{farhadi2009describing}
A.~Farhadi, I.~Endres, D.~Hoiem, and D.~Forsyth.
\newblock Describing objects by their attributes.
\newblock In {\em CVPR}, 2009.

\bibitem{felzenszwalb2010object}
P.~F. Felzenszwalb, R.~B. Girshick, D.~McAllester, and D.~Ramanan.
\newblock Object detection with discriminatively trained part-based models.
\newblock {\em IEEE transactions on pattern analysis and machine intelligence},
  32(9):1627--1645, 2010.

\bibitem{frome2013devise}
A.~Frome, G.~S. Corrado, J.~Shlens, S.~Bengio, J.~Dean, T.~Mikolov, et~al.
\newblock Devise: A deep visual-semantic embedding model.
\newblock In {\em NIPS}, 2013.

\bibitem{galleguillos2008object}
C.~Galleguillos, A.~Rabinovich, and S.~Belongie.
\newblock Object categorization using co-occurrence, location and appearance.
\newblock In {\em CVPR}, 2008.

\bibitem{garcia2018few}
V.~Garcia and J.~Bruna.
\newblock Few-shot learning with graph neural networks.
\newblock In {\em ICLR}, 2018.

\bibitem{NIPS2009_3766}
S.~Gould, T.~Gao, and D.~Koller.
\newblock Region-based segmentation and object detection.
\newblock In {\em Advances in Neural Information Processing Systems 22}, 2009.

\bibitem{he2016deep}
K.~He, X.~Zhang, S.~Ren, and J.~Sun.
\newblock Deep residual learning for image recognition.
\newblock In {\em CVPR}, 2016.

\bibitem{hoffman2014lsda}
J.~Hoffman, S.~Guadarrama, E.~S. Tzeng, R.~Hu, J.~Donahue, R.~Girshick,
  T.~Darrell, and K.~Saenko.
\newblock Lsda: Large scale detection through adaptation.
\newblock In {\em NIPS}, 2014.

\bibitem{hu2017relation}
H.~Hu, J.~Gu, Z.~Zhang, J.~Dai, and Y.~Wei.
\newblock Relation networks for object detection.
\newblock In {\em CVPR}, 2018.

\bibitem{hu2018learning}
R.~Hu, P.~Dollár, K.~He, T.~Darrell, and R.~Girshick.
\newblock Learning to segment every thing.
\newblock In {\em CVPR}, 2018.

\bibitem{jayaraman2014zero}
D.~Jayaraman and K.~Grauman.
\newblock Zero-shot recognition with unreliable attributes.
\newblock In {\em NIPS}, 2014.

\bibitem{kipf2016semi}
T.~N. Kipf and M.~Welling.
\newblock Semi-supervised classification with graph convolutional networks.
\newblock In {\em ICLR}, 2017.

\bibitem{krishnavisualgenome}
R.~Krishna, Y.~Zhu, O.~Groth, J.~Johnson, K.~Hata, J.~Kravitz, S.~Chen,
  Y.~Kalantidis, L.-J. Li, D.~A. Shamma, M.~Bernstein, and L.~Fei-Fei.
\newblock Visual genome: Connecting language and vision using crowdsourced
  dense image annotations.
\newblock {\em International Journal of Computer Vision}, 2016.

\bibitem{lampert2009learning}
C.~H. Lampert, H.~Nickisch, and S.~Harmeling.
\newblock Learning to detect unseen object classes by between-class attribute
  transfer.
\newblock In {\em CVPR}, 2009.

\bibitem{li2017attentive}
J.~Li, Y.~Wei, X.~Liang, J.~Dong, T.~Xu, J.~Feng, and S.~Yan.
\newblock Attentive contexts for object detection.
\newblock {\em IEEE Transactions on Multimedia}, 19(5):944--954, 2017.

\bibitem{li2015gated}
Y.~Li, D.~Tarlow, M.~Brockschmidt, and R.~Zemel.
\newblock Gated graph sequence neural networks.
\newblock In {\em ICLR}, 2016.

\bibitem{li2017light}
Z.~Li, C.~Peng, G.~Yu, X.~Zhang, Y.~Deng, and J.~Sun.
\newblock Light-head r-cnn: In defense of two-stage object detector.
\newblock In {\em CVPR}, 2017.

\bibitem{lin2017feature}
T.-Y. Lin, P.~Doll{\'a}r, R.~Girshick, K.~He, B.~Hariharan, and S.~Belongie.
\newblock Feature pyramid networks for object detection.
\newblock In {\em CVPR}, 2017.

\bibitem{lin2014microsoft}
T.-Y. Lin, M.~Maire, S.~Belongie, J.~Hays, P.~Perona, D.~Ramanan,
  P.~Doll{\'a}r, and C.~L. Zitnick.
\newblock Microsoft coco: Common objects in context.
\newblock In {\em ECCV}, 2014.

\bibitem{liu2016ssd}
W.~Liu, D.~Anguelov, D.~Erhan, C.~Szegedy, S.~Reed, C.-Y. Fu, and A.~C. Berg.
\newblock Ssd: Single shot multibox detector.
\newblock In {\em ECCV}, 2016.

\bibitem{maaten2008visualizing}
L.~v.~d. Maaten and G.~Hinton.
\newblock Visualizing data using t-sne.
\newblock {\em Journal of machine learning research}, 9(Nov):2579--2605, 2008.

\bibitem{mao2015learning}
J.~Mao, X.~Wei, Y.~Yang, J.~Wang, Z.~Huang, and A.~L. Yuille.
\newblock Learning like a child: Fast novel visual concept learning from
  sentence descriptions of images.
\newblock In {\em ICCV}, 2015.

\bibitem{marino2016more}
K.~Marino, R.~Salakhutdinov, and A.~Gupta.
\newblock The more you know: Using knowledge graphs for image classification.
\newblock In {\em CVPR}, 2017.

\bibitem{mensink2012metric}
T.~Mensink, J.~Verbeek, F.~Perronnin, and G.~Csurka.
\newblock Metric learning for large scale image classification: Generalizing to
  new classes at near-zero cost.
\newblock In {\em Computer Vision--ECCV 2012}, 2012.

\bibitem{miller1995wordnet}
G.~A. Miller.
\newblock Wordnet: a lexical database for english.
\newblock {\em Communications of the ACM}, 38(11):39--41, 1995.

\bibitem{misra2017red}
I.~Misra, A.~Gupta, and M.~Hebert.
\newblock From red wine to red tomato: Composition with context.
\newblock In {\em CVPR}, 2017.

\bibitem{mottaghi2014role}
R.~Mottaghi, X.~Chen, X.~Liu, N.-G. Cho, S.-W. Lee, S.~Fidler, R.~Urtasun, and
  A.~Yuille.
\newblock The role of context for object detection and semantic segmentation in
  the wild.
\newblock In {\em CVPR}, 2014.

\bibitem{niepert2016learning}
M.~Niepert, M.~Ahmed, and K.~Kutzkov.
\newblock Learning convolutional neural networks for graphs.
\newblock In {\em ICML}, pages 2014--2023, 2016.

\bibitem{parikh2011relative}
D.~Parikh and K.~Grauman.
\newblock Relative attributes.
\newblock In {\em ICCV}, 2011.

\bibitem{paszke2017automatic}
A.~Paszke, S.~Gross, S.~Chintala, G.~Chanan, E.~Yang, Z.~DeVito, Z.~Lin,
  A.~Desmaison, L.~Antiga, and A.~Lerer.
\newblock Automatic differentiation in pytorch.
\newblock In {\em NIPS Workshop}, 2017.

\bibitem{redmon2016you}
J.~Redmon, S.~Divvala, R.~Girshick, and A.~Farhadi.
\newblock You only look once: Unified, real-time object detection.
\newblock In {\em CVPR}, 2016.

\bibitem{reed2016learning}
S.~Reed, Z.~Akata, H.~Lee, and B.~Schiele.
\newblock Learning deep representations of fine-grained visual descriptions.
\newblock In {\em CVPR}, 2016.

\bibitem{ren2015faster}
S.~Ren, K.~He, R.~Girshick, and J.~Sun.
\newblock Faster r-cnn: Towards real-time object detection with region proposal
  networks.
\newblock In {\em NIPS}, 2015.

\bibitem{rohrbach2013transfer}
M.~Rohrbach, S.~Ebert, and B.~Schiele.
\newblock Transfer learning in a transductive setting.
\newblock In {\em NIPS}, 2013.

\bibitem{russakovsky2015imagenet}
O.~Russakovsky, J.~Deng, H.~Su, J.~Krause, S.~Satheesh, S.~Ma, Z.~Huang,
  A.~Karpathy, A.~Khosla, M.~Bernstein, et~al.
\newblock Imagenet large scale visual recognition challenge.
\newblock {\em International Journal of Computer Vision}, 115(3):211--252,
  2015.

\bibitem{salakhutdinov2011learning}
R.~Salakhutdinov, A.~Torralba, and J.~Tenenbaum.
\newblock Learning to share visual appearance for multiclass object detection.
\newblock In {\em CVPR}, 2011.

\bibitem{simonyan2014very}
K.~Simonyan and A.~Zisserman.
\newblock Very deep convolutional networks for large-scale image recognition.
\newblock In {\em ICLR}, 2015.

\bibitem{springenberg2014striving}
J.~T. Springenberg, A.~Dosovitskiy, T.~Brox, and M.~Riedmiller.
\newblock Striving for simplicity: The all convolutional net.
\newblock In {\em ICLR Workshop}, 2015.

\bibitem{torralba2004sharing}
A.~Torralba, K.~P. Murphy, and W.~T. Freeman.
\newblock Sharing features: efficient boosting procedures for multiclass object
  detection.
\newblock In {\em CVPR}, 2004.

\bibitem{torralba2003context}
A.~Torralba, K.~P. Murphy, W.~T. Freeman, and M.~A. Rubin.
\newblock Context-based vision system for place and object recognition.
\newblock In {\em ICCV}, 2003.

\bibitem{Vaswani2017Attention}
A.~Vaswani, N.~Shazeer, N.~Parmar, J.~Uszkoreit, L.~Jones, A.~N. Gomez,
  L.~Kaiser, and I.~Polosukhin.
\newblock Attention is all you need.
\newblock In {\em NIPS}, 2017.

\bibitem{Wang2017Non}
X.~Wang, R.~Girshick, A.~Gupta, and K.~He.
\newblock Non-local neural networks.
\newblock In {\em CVPR}, 2018.

\bibitem{wang2018zero}
X.~Wang, Y.~Ye, and A.~Gupta.
\newblock Zero-shot recognition via semantic embeddings and knowledge graphs.
\newblock In {\em CVPR}, 2018.

\bibitem{wu2016ask}
Q.~Wu, P.~Wang, C.~Shen, A.~Dick, and A.~van~den Hengel.
\newblock Ask me anything: Free-form visual question answering based on
  knowledge from external sources.
\newblock In {\em CVPR}, 2016.

\bibitem{jjfaster2rcnn}
J.~Yang, J.~Lu, D.~Batra, and D.~Parikh.
\newblock A faster pytorch implementation of faster r-cnn.
\newblock {\em https://github.com/jwyang/faster-rcnn.pytorch}, 2017.

\bibitem{zhou2017scene}
B.~Zhou, H.~Zhao, X.~Puig, S.~Fidler, A.~Barriuso, and A.~Torralba.
\newblock Scene parsing through ade20k dataset.
\newblock In {\em CVPR}, 2017.

\end{thebibliography}

\end{document}